\documentclass[letterpaper, 10 pt, conference]{ieeeconf}  % Comment this line out if you need a4paper

\IEEEoverridecommandlockouts                              % This command is only needed if 
\newcounter{subeq}

\usepackage{mdframed}
\usepackage{colonequals}
\usepackage{amsmath,amsfonts,amssymb}
\usepackage{graphicx}
\usepackage{microtype}
\usepackage{xspace}
\usepackage{subfigure}
\usepackage{booktabs}
\usepackage{nicefrac}
\usepackage{paralist}
\usepackage{tikz}
\usepackage{multirow}
\usepackage{algorithm2e,theorem,balance}
\usetikzlibrary{arrows,decorations.pathmorphing,positioning,fit,trees,shapes,shadows,automata,calc,backgrounds} 
\tikzset{>=stealth}
\tikzset{state/.append style={inner sep=2pt,minimum size=2pt}}

\newtheorem{definition}{Definition}

\newtheorem{theorem}{Theorem}
\newtheorem{example}{Example}

\newtheorem{proposition}{Proposition}

\usepackage[utf8]{inputenc}
\usepackage{tabularx,ragged2e,booktabs}

\usepackage{colonequals}
\usepackage{amsmath,amsfonts,amssymb}
\usepackage{graphicx}
\usepackage{microtype}
\usepackage{xspace}
\usepackage{booktabs}
\usepackage{nicefrac}
\usepackage{paralist}
\usepackage{tikz}
\usepackage{multirow}
\usepackage{algorithm2e,theorem,balance}
\usetikzlibrary{arrows,decorations.pathmorphing,positioning,fit,trees,shapes,shadows,automata,calc,backgrounds} 
\tikzset{>=stealth}
\tikzset{state/.append style={inner sep=2pt,minimum size=2pt}}

\title{\LARGE \bf
The Partially Observable Games We Play for Cyber Deception
}

\author{Mohamadreza Ahmadi, Murat Cubuktepe, Nils Jansen, Sebastian Junges \\Joost-Pieter Katoen and Ufuk Topcu% <-this % stops a space
\thanks{M. Ahmadi, M. Cubuktepe, and U. Topcu are with the Department of Aerospace Engineering and Engineering Mechanics, University of Texas, 201 E 24th St, Austin, TX 78712, USA. N. Jansen is with the Department of Software Science, Radboud University Nijmegen, Comeniuslaan 4, 6525 HP Nijmegen, Netherlands. S. Junges and J-P. Katoen are with the Department of Computer Science, RWTH Aachen University, Templergraben 55, 52062 Aachen, Germany. e-mail: (\{mrahmadi,mcubuktepe, utopcu\}@utexas.edu,n.jansen@science.ru.nl,\{sebastian.junges, katoen\}@cs.rwth-aachen.de). 
}%
}

\newtheorem{problem}{Task }
\newcommand{\ie}{i.\,e.\@\xspace}

\newcommand{\p}{\ensuremath{\mathbb{P}}}
\newcommand{\pr}{\ensuremath{\mathrm{Pr}}}

\newcommand{\reachProp}[2]{\ensuremath{\p_{\leq #1}(\finally #2)}}

\newcommand{\er}{\ensuremath{\mathrm{EC}}}

\newcommand{\expRewProp}[2]{\ensuremath{\er_{\leq #1}(\finally #2)}}

\newcommand{\finally}{\lozenge}
\newcommand{\R}{\mathbb{R}}

\newcommand{\N}{\mathbb{N}}

\newcommand{\FSCs}[2][]{\ensuremath{\mathit{FSC}_{#1}^{#2}}}

\newcommand{\Ireal}{[0,\, 1]\subseteq\mathbb{R}}  % real interval [0,1]
\newcommand{\Distr}{\mathit{Distr}}

\newcommand{\distDom}{X}

\newcommand{\distFunc}{\mu}
\newcommand{\distDomElem}{x}
\DeclareMathOperator{\supp}{supp}

\DeclareMathOperator{\dom}{dom}

\newcommand{\Var}{\ensuremath{V}\xspace}        % Set of program variables 
\newcommand{\sinit}{s_{\mathrm{I}}} % initial state of DTMC/MDP

\newcommand{\ninit}{n_{\mathrm{I}}}
\newcommand{\mdp}{M}

\newcommand{\pMdpInit}[1][]{\ensuremath{\mdp{#1}=(S{#1},\,\sinit{#1},\Act{#1},\Var,\probmdp{#1})}}
\newcommand{\probmdp}{\mathcal{P}}

\newcommand{\fsc}{\ensuremath{\mathcal{A}}}

\newcommand{\actionMap}{\ensuremath{\mathcal{\gamma}}}
\newcommand{\nodeTransition}{\ensuremath{\mathcal{\delta}}}
\newcommand{\FSCinit}{\fsc=(N,\ninit,\actionMap,\nodeTransition)}
\newcommand{\ObsSym}{{Z}}
\newcommand{\ObsFun}{{O}}
\newcommand{\obs}{\ensuremath{z}}

\newcommand{\pomdp}{\mathcal{M}}
\newcommand{\posg}{\mathcal{G}}
\newcommand{\PosgInit}[1][]{\posg{#1}=(\sg{#1},\ObsSym{#1},\ObsFun{#1})}

\newcommand{\spOne}{\ensuremath{S_{\circ}}}
\newcommand{\spTwo}{\ensuremath{S_{\Box}}}

\newcommand{\probsg}{\probmdp}
\newcommand{\sg}{\ensuremath{G}}  %% changed back to G to distinguish from MDP
\newcommand{\sgInit}[1][]{\ensuremath{\sg{#1}=(\spOne{#1},\spTwo{#1},\sinit{#1},\Act,\probsg{#1})}}

\newcommand{\states}{\ensuremath{S}}

\newcommand{\poly}[1][]{\ensuremath{\mathbb{Q}[#1]}}

\newcommand{\sched}{\ensuremath{\sigma}}
\newcommand{\Sched}{\ensuremath{{\Sigma}}}
\newcommand{\osched}{\ensuremath{\mathit{\sigma}}}

\newcommand{\Act}{\ensuremath{\mathit{Act}}}
\newcommand{\act}{\ensuremath{a}}
\newcommand{\pmdp}{\ensuremath{M}}

\newcommand{\pathset}{\mathsf{Paths}}

\newcommand{\pathsfin}{\pathset_{\mathit{fin}}}

\newcommand{\last}[1]{\mathrm{last}(#1)}

\newcommand{\IStrat}{\mathsf{IStrat}}
\newcommand{\infiltration}{\mathcal{I}}

\DeclareMathAlphabet{\mathpzc}{OT1}{pzc}{m}{it}
\def\presuper#1#2%
  {\mathop{}%
   \mathopen{\vphantom{#2}}^{#1}%
   \kern-\scriptspace%
   #2}

\begin{document}

%\emptypage
\maketitle
\begin{abstract}
Progressively intricate cyber infiltration mechanisms have made conventional means of defense, such as firewalls and malware detectors, incompetent. 
These sophisticated infiltration mechanisms can study the defender's behavior, identify security caveats, and modify their actions adaptively. To tackle these security challenges, cyber-infrastructures require active defense techniques that incorporate cyber deception, in which the defender (deceiver)  implements a strategy to mislead the infiltrator. To this end, we use a two-player partially observable stochastic game (POSG) framework, wherein the deceiver has full observability over the states of the POSG, and the infiltrator has partial observability. Then, the deception problem is to  compute a strategy for the deceiver that minimizes the expected cost of deception against all strategies of the infiltrator. We first show that the underlying problem is a robust mixed-integer linear program, which is intractable to solve in general. Towards a scalable approach, we compute optimal finite-memory strategies for the infiltrator by a reduction to a series of synthesis problems for parametric Markov decision processes. We use these infiltration strategies to find robust strategies for the deceiver using mixed-integer linear programming. We illustrate the performance of our technique on a POSG model for network security. Our experiments demonstrate that the proposed approach handles scenarios considerably larger than those of the state-of-the-art methods. 
\end{abstract}

\section{Introduction}
\label{sec:introduction}

``If deception\footnote{Deceiving means to ``deliberately cause (someone) to believe something that is not true, especially for personal gain'' according to the Oxford English Dictionary.} can be used for cyber-attacks, can it also be used for defense?''. This question was raised by the renowned hacker, Kevin Mitnick, in his famous book The Art of Deception~\cite{mitnick2011art}. The reason for asking for such advanced cyber-defense mechanisms is the ever-escalating progress of cyber-infiltration~\cite{tankard2011advanced}. Furthermore, the prolific reliance of governments, industries, and individuals on cyber-infrastructure, thanks to the growth of applications of AI and machine learning tools, makes them particularly attractive for cyber-terrorism and cyber-crime~\cite{data2014privacy}. 

Deception is a familiar technique  for war zone strategists~\cite{daniel2013strategic} and for cyber-infiltrators or hackers~\cite{denning1999information}. In cyber-infiltration, for example, deception can be achieved by changing  malware signatures, social engineering, concealing codes, and encrypting exploits. On the other hand, deception defense strategies can use deceits and feints
to thwart an infiltrator's cognitive processes, disrupt the breach process, and delay
infiltration activities. Such deception can be carried out via misleading, obfuscations, and fake
responses. These methods rely on the infiltrator's \emph{belief} in the network responses and data. For instance, honeypot servers (fake servers that mimic actual servers)  are commonly used to actively detect malicious activity and reveal the infiltrator's strategies~\cite{spitzner2003honeypots}. From a cyber-deception standpoint, two factors, namely, the amount of deception and the frequency of deception, characterize the cost of implementing a deception mechanism. 

Cyber-deception can be mathematically formalized as a non-cooperative two-player dynamic game~\cite{basar1999dynamic,manshaei2013game,zhu2015game}, which can represent  the adversarial sequential decision making nature of a deception/infiltration scenario and the limited cyber-defense/infiltration resources. In particular, games with imperfect information can represent the information asymmetry, which is at the heart of cyber-deception, i.e., the deceiver has full knowledge over the cyber-infrastructure, such as servers, but the infiltrator does not. Recently, in~\cite{horak2017manipulating}, the authors proposed a modeling framework for cyber-deception in terms of a one-sided partially observable Markov decision processes-a special class of partially observable Markov decision processes (POSGs), where one player has full observability and the second one does not. Despite this unique modeling paradigm, POSGs are notoriously intractable to solve in general~\cite{hansen2004dynamic}. Three  approximate methods  have been proposed in the literature for solving POSGs either by a memory bounded representation of the value function
~\cite{emery2004approximate}, by approximating it by a series of smaller, related Bayesian games using heuristics \cite{kumar2009dynamic}, or by using heuristic search value iteration~\cite{horak2017heuristic}. 

For POSGs, the optimal strategies for the deceiver may require infinite-memory, or they may require unbounded-memory. To obtain a tractable formulation, we use finite-memory strategies for the infiltrator. Using finite memory, we have a compact representation of the infiltrator strategies. Then, we search for strong finite-memory infiltrator strategies that induces a low cost for the infiltrator. The set of all finite-memory infiltrator strategies is uncountable, and the measure of the finite-memory strategies that induces a low cost for the infiltrator may be small compared to the set of all finite-memory strategies. Therefore, methods such as evolutionary algorithms or Bayesian methods may not be applicable in this setting. We define the infiltrator strategy synthesis problem as a synthesis problem in parametric MDPs~\cite{hahn2011synthesis}, and we use our previous work~\cite{cubuktepe-et-al-qcqp-techreport} to synthesize strategies in a parametric MDP using convex-concave approach~\cite{lipp2016variations}. Finally, using the infiltrator strategies, we compute a robust deceiver strategy that maximizes the worst-case cost of the all strategies using mixed-integer linear programming (MILP). We demonstrate that, we can compute strategies for a deception game with higher number of states and action than those that can be tackled by  the state-of-the-art methods.

The rest of the paper is organized as follows. In the next section, we motivate the problem studied in this paper with a motivating example. In Section III, we provide some preliminary definitions and formulate the problem. In Section IV, we describe our two-stage tractable approach. In Section V, we apply the proposed approach to the motivating example. Finally, we conclude the paper in Section VI and give directions for future research.
\section{Motivating Example: \\Active Deception for Network Security}\label{sec:example}

\begin{figure}[!t]\label{fig:network}
\centering
\includegraphics[scale=0.42]{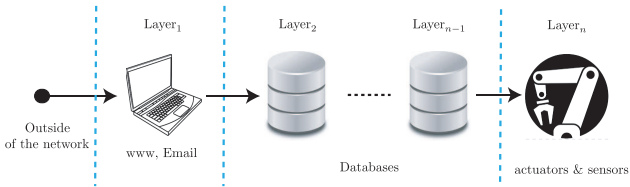}
\caption{Multi-layer network topology.}
\end{figure}

\begin{figure*}[!t]\label{fig:networksecurity}
\centering
\includegraphics[scale=0.5]{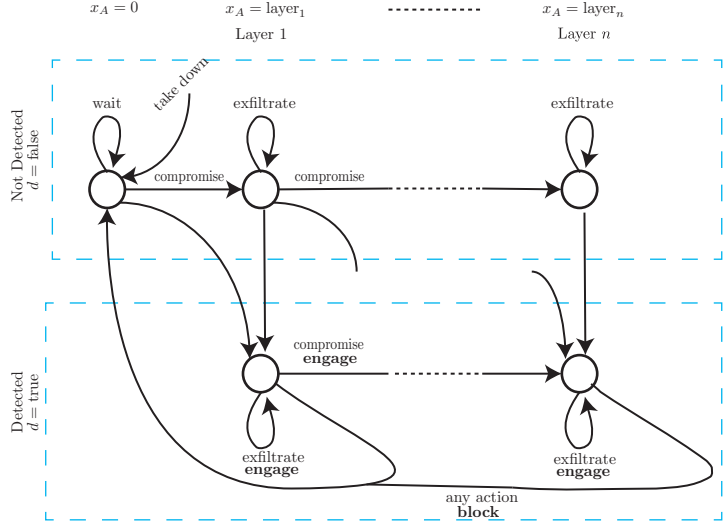}
\caption{Schematic diagram of the transition system of the one-sided POSG model for network security. The bold actions correspond to the deceiver; whereas, the rest belong to the infiltrator.}
\end{figure*}

We motivate the problem under study in this paper by a computer network security example adapted from~\cite{horak2017manipulating}. Consider a multilayer network used typically in critical network operations such as power plants and manufacturing facilities~\cite{kuipers2006control,byres2004myths}~(see Figure~1). The infiltrator starts the attack from outside of the network and proceeds by accessing deeper layers, where more sensitive/valuable assets are located. The middle layers are composed of databases containing confidential data and the last layer provides access to physical devices. Infiltration to the last layer of the network can lead to critical damage to the facility~\cite{kushner2013real}.

The network we consider consists of $n$ layers as illustrated in Figure~2. Each column in Figure~2 corresponds to a network layer, e.g., emails, databases, actuators and sensors, and etc. 

The deceiver's task here is to manipulate the infiltrator's belief over whether he/she is being detected. The deceived infiltrator is hence convinced to take wrong actions due to the uncertainty about the infiltration progress. The deceiver has to manipulate the infiltrator's belief and keep him $engage$d in the network.  

One way to represent the network security problem depicted in Figure~2 is to represent it by a \emph{one-sided POSG} (to be described formally in the sequel), which  is comprised of two parts. The upper-half corresponds to the states, wherein the presence of the infiltrator in the network has not been detected ($d = \mathrm{false}$) and the deception mechanism cannot be implemented; whereas, the lower half corresponds to the case wherein the infiltrator is detected and therefore we can choose either actions $``\mathrm{engage}"$ or $``\mathrm{block}"$. 

The arrows illustrate the transitions in the game characterized by a transition function $T$. We assume all transitions are deterministic, except transitions from $d=\mathrm{false}$ to $d=\mathrm{true}$. 

The infiltrator starts the attack at a computer outside the network $x_A=0$. The infiltrator attempts to access computers in deeper layers by $compromise$ing them. He/she also has the option to $wait$ at any layer, $take down$, which reveals the presence of the infiltrator and the infiltrator is forced to exit the network and restart the attack. Another course of action is to incur small amount of damage by $exfiltrate$ing of data, which does not attract the deceiver's attention.

The infiltrator does not receive observations for detection state $d$, but is aware of $x_A$ (the network  Layer being infiltrated). The deceiver has no information over the infiltrator until the infiltrator has been detected.

\section{Problem Statement}
In this section, we first lay the formal foundations for POSGs, followed by a formal problem statement together with the underlying optimization problem.

\subsection{Foundations}
A \emph{probability distribution} over a finite or countably infinite set $\distDom$
is a function $\distFunc\colon\distDom\rightarrow\Ireal$ with $\sum_{\distDomElem\in\distDom}\distFunc(\distDomElem)=\distFunc(\distDom)=1$.
The set of all distributions on $\distDom$ is $\Distr(\distDom)$. The support of a distribution $\distFunc$ is
$\supp(\distFunc) = \{x\in\distDom\,|\,\distFunc(x)>0\}$.
A distribution is \emph{Dirac} if $|\!\supp(\distFunc)| = 1$. Let $V=\{x_1,\ldots,x_n\}$ be a finite set of \emph{variables} over the real numbers $\R$. The set of multivariate polynomials over $V$ is $\mathbb{Q}[V]$. An \emph{instantiation} for $V$ is a function $u\colon V \rightarrow \R$.

\begin{definition}[SG]
  \label{def:sg}
  A \emph{stochastic game} (SG) is a tuple $\sgInit$ with a finite set $\states=\spOne\cup\spTwo$ of \emph{states}, a set $\spOne$ of Player~1 states, a set $\spTwo$ of Player~2 state, the \emph{initial state} $\sinit\in\states$, a finite set $\Act$ of \emph{actions}, and a \emph{transition function} 
  $\probmdp\colon \states\times\Act\rightarrow\Distr(\states)$. We define costs using a state-action cost function $C : S \times \Act \rightarrow \mathbb{R}_{\geq 0}$.
\end{definition}
%%\sj{If we speak about rewards, they should be defined somewhere}
%

A Markov decision process (MDP) is an SG in which $\spOne = \emptyset$, and consequently $S = \spTwo$. A \emph{path} of an SG $\sg$ is an (in)finite sequence $\pi = s_0\xrightarrow{\act_0}s_1\xrightarrow{\act_1} ~s$,
where $s_0=\sinit$, $s_i\in\states$, $\act_i\in\Act$, and $\probmdp(s_i,\act_i)(s_{i+1})\neq 0$ for all $i\in\N$.
For finite $\pi$, $\last{\pi}$ denotes the last state of $\pi$. 
The set of (in)finite paths of $\sg$ is $\pathsfin^{\sg}$ ($\pathset^{\sg}$).

%To define a probability measure over the paths of a PG $\pg$, the non-determinism needs to be
%resolved by \emph{strategies}.
\begin{definition}[SG strategy]
  \label{def:strategy}
  A \emph{strategy} $\sched$ for $\sg$ is a pair $\sched = (\sched_1,\sched_2)$ of functions
  $\sched_i\colon \{\pi\in \pathsfin^{\sg} \mid\last{\pi}\in S_i\}\to\Distr(\Act)$ such that for all $\pi\in \pathsfin^{\sg}$,
  $\{ \act\mid\sched_i(\pi)(\act)>0\} \subseteq \Act\bigl(\last{\pi}\bigr)$. 
%  $\scheds[\pg]$ denotes the set of all strategies of $\pg$
%  and $\scheds[\pg]^i$ all Player-$i$ strategies of $\pg$.
\end{definition}
A Player-$i$ strategy $\sched_i$ (for $i\in\{1,2\}$) is \emph{memoryless} if $\last{\pi}=\last{\pi'}$ implies $\sched_i(\pi)=\sched_i(\pi')$ for all $\pi,\pi'\in\dom(\sched_i)$. 
It is \emph{deterministic} if $\sched_i(\pi)$ is a Dirac distribution for all $\pi\in\dom(\sched_i)$. 
%A \emph{memoryless deterministic} strategy is of the form $\sched_i\colon S_i\to \acts$.

%A strategy~$\sigma$ for a PG resolves all non-deterministic choices, yielding an \emph{induced MC}, for which a \emph{probability measure} over the set of infinite paths is defined by the standard cylinder set construction~\cite{BK08}.
%These notions are analogous for MDPs.

\begin{definition}[One-sided POSG]
  \label{def:pomdp}
  A \emph{one-sided partially observable stochastic game (POSG)} is a tuple $\PosgInit$, with $\sgInit$ the \emph{underlying SG of $\posg$}, $\ObsSym$ a finite set of observations, and $\ObsFun\colon S\rightarrow\ObsSym$ the \emph{observation function} for Player~2.
\end{definition}

A partially observable Markov decision process (POMDP) is an one-sided POSG in which $\spOne = \emptyset$, and consequently $S = \spTwo$. In a POSG, players cannot make any choice when $|\Act(s)| = 1$. Thus, for a POSG $\posg$ with states $\spOne \cup \spTwo$, the POSG $\posg'$ with states $\spOne' = \spOne \setminus \{ s \in \spOne \mid |\Act(s)| = 1 \}$ and $\spTwo' = \spTwo \cup  \{ s \in \spOne \mid |\Act(s)| = 1 \}$ and all transitions unchanged is equivalent w.r.t. the properties considered in the paper.
Consequently, a POSG where one player never has a choice, i.e. $|\Act(s)| = 1,~\forall s \in \spOne$ is also an MDP.

Without loss of generality, we assume Player~2 can observe its own available actions, thus $|\Act(s)| = |\Act(s')|$ for all $s, s' \in \spTwo$ with $\ObsFun(s) = \ObsFun(s')$.
Essentially, in a one-sided POSG, Player~1 has full observability, while Player~2 has only partial observability.
We lift the observation function to paths: For $\pi=s_0\xrightarrow{\act_0} s_1\xrightarrow{\act_1} ~s s_n\in\pathsfin^{\mdp}$, the associated \emph{observation sequence} is $\ObsFun(\pi)=\ObsFun(s_0)\xrightarrow{\act_0} \ObsFun(s_1)\xrightarrow{\act_1} ~s\ObsFun(s_n)$.
%Several paths in the underlying MDP may yield the same observation sequence.
%Strategies have to take this restricted observability into account.

\begin{definition}[POSG Strategy]
  \label{def:obsstrategy}
  An \emph{observation-based strategy} $\osched=(\osched_1,\osched_2)$ for a one-sided POSG $\posg$ is a strategy $(\osched_1,\osched_2)$ for the underlying SG $\sg$ such that $\osched_2(\pi)=\osched_2(\pi')$ for all $\pi,\pi'\in\pathsfin^{\posg}$
  with $\ObsFun(\pi)=\ObsFun(\pi')$. 
  $\osched_1$ is the Player~1 strategy, and $\osched_2$ is the (observation-based) Player~2 strategy.
%  \nj{does it make sense like this?}  
%  $\osched_1\colon S\rightarrow\Distr(\Act)$ the Player~1 strategy, and $\osched_2\colon S\rightarrow\Distr(\Act)$ and \nj{continue here}

%  $\osched$ for the underlying 
%  such that $\osched(\pi)=\osched(\pi')$ for all $\pi,\pi'\in\pathsfin^{\mdp}$
%  with $\ObsFun(\pi)=\ObsFun(\pi')$.
%  $\oSched^\pomdp$ is the set of observation-based strategies for $\pomdp$.
\end{definition}
Applying the strategy $\osched=(\osched_1,\osched_2)$ to a POSG $\posg$ resolves all nondeterminism and partial observability, resulting in the \emph{induced Markov chain} $\posg^\osched$.
%For an MDP $\mdp$, there is a memoryless deterministic strategy inducing the maximal (or minimal) probability $\pr_{\max}^\mdp(\neg B\,\Until\,G)$~\cite{Condon92}.
In general, POSGs extend POMDPs and optimal strategies require infinite memory in general~\cite{chatterjee2016decidable}.
To represent observation-based strategies with \emph{finite} memory, we use \emph{finite-state controllers} (FSCs).  If such an FSC has $n$ memory states, we speak of memory size $n$ for the underlying strategy $\osched$.

%The problem of proving the satisfaction of $\varphi$ is therefore undecidable~\cite{ChatterjeeCT16}.
\begin{definition}[FSC]
  \label{def:fsc}
  A \emph{finite-state controller (FSC)} for a POMDP $\pomdp$ is a tuple $\FSCinit$, where $N$ is a finite set of \emph{memory nodes},
  $\ninit\in N$ is the \emph{initial memory node}, $\actionMap$ is the \emph{action mapping} $\actionMap\colon N\times\ObsSym\rightarrow\Distr(\Act)$,
  and $\nodeTransition$ is the \emph{memory update} $\nodeTransition\colon N\times\ObsSym\times\Act\rightarrow \Distr(N)$.
  The set $\FSCs[k]{\pomdp}$ denotes the set of FSCs with $k$ memory nodes, called \emph{$k$-FSC}s.
%  Let $\osched_\fsc\in\oSched^\pomdp$ denote the observation-based strategy represented by $\fsc$.
\end{definition}
From a node $n$ and the observation $\obs$ in the current state of the POMDP, the next action $\act$ is
 chosen from $\Act(\obs)$ randomly as given by $\actionMap(n, \obs)$. Then, the successor node of the FSC is determined randomly via $\nodeTransition(n, \obs, \act)$.

\emph{Specifications.}
For a POSG $\posg$ and a set $T\subseteq\states$ of \emph{target states}, we consider the maximal (or minimal) \emph{probability} $\pr_{\max}^\posg(\finally T)$ ($\pr_{\min}^\posg(\finally T)$) to reach $T$, as well as the maximal (or minimal) \emph{expected cost} $\er_{\max}^\posg(\finally T)$ ($\er_{\min}^\posg(\finally T)$).
For a probability bound $\lambda\in[0,1]$ and an expected cost bound $\kappa\in\R$, we also consider specifications of the form $\varphi=\reachProp{\lambda}{T}$ and $\psi=\expRewProp{\kappa}{T}$, where the probability or the expected cost to reach $T$ shall be at most $\lambda$ or $\kappa$, respectively.
The specification $\varphi$ is satisfied for a strategy $\osched=(\osched_1,\osched_2)$ and the POSG $\posg$ if the probability $\pr^{\posg^\osched}(\finally T)$ of reaching a target state in $\posg^\osched$ is at most $\lambda$, denoted by $\pomdp^\osched\models\varphi$.
This satisfaction relation is analogous for expected cost.

\emph{Sufficiently strong strategies.} 
For a POSG $\posg$ and a property $\varphi$, a strategy $\sigma_1$ for Player 1 is called \emph{sufficiently strong} for $\varphi$ against a set $\mathcal{S} \subseteq \Sched_2$ of strategies for Player 2, if for each strategy $\sigma_2 \in \mathcal{S}$ of Player 2, $\posg^{(\sigma_1,\sigma_2)} \models \varphi$.

%, and a \emph{threshold} $\lambda \in [0,1]$, we consider \emph{quantitative reachability specifications}
%$\varphi=\p_{\leq \lambda} (\finally B)$. 
%The specification $\varphi$ is satisfied for a strategy $\osched\in\oSched^\pomdp$ if the probability $\pr^{\pomdp^\osched}(\finally B)$ of reaching a bad state in $\pomdp^\osched$ is at most $\lambda$, denoted by $ \pomdp^\osched\models\varphi$.
%Vice versa, a strategy $\sched'\in\oSched^\pomdp$ refutes the specification, denoted by $\pomdp^\osched\not\models\varphi$, if $\pr^{\pomdp^\osched}(\finally B)>\lambda$.
%
%
%For an MDP $\mdp$, there is a memoryless deterministic strategy inducing the maximal (or minimal) probability $\pr_{\max}^\mdp(\neg B\,\Until\,G)$~\cite{Condon92}.
%For a POMDP $\pomdp$, however, observation-based strategies with infinite memory as in Def.~\ref{def:obsstrategy} are necessary~\cite{Ross83}.
%The problem of proving the satisfaction of $\varphi$ is therefore undecidable~\cite{ChatterjeeCT16}.

\subsection{Deception Games}

\begin{definition}[Deception Game]
A \emph{deception game} is a one-sided POSG, where Player~1 (with full observability) is called the \emph{infiltrator} and Player~2 (with partial observability) is called the \emph{deceiver}. See Figure~\ref{fig:flowchart_problem} for an illustration of the robust deception problem.
\end{definition}
%\vspace{-4cm}
%\nj{Reza: draw concrete relation to example here.}
In the POSG that captures the motivating example, we identify the target states $T\subseteq S$ as the states that correspond to Layer~$n$.
The goal is to \emph{deceive} the infiltrator such that the expected cost to reach those states is maximized. 
We state the formal problem.

\begin{mdframed}[backgroundcolor=gray!50!white]\label{prob1}
Problem~1: Given a deception game $\PosgInit$, a set of target states $T\subseteq S$, memory bound $n$, and  specification $\varphi = \er_{\min}^\posg(\finally T)$, compute a sufficiently strong strategy for the deceiver, against infiltration-strategies  with memory $n'\leq n$.
\end{mdframed}
%\nj{improve}

{
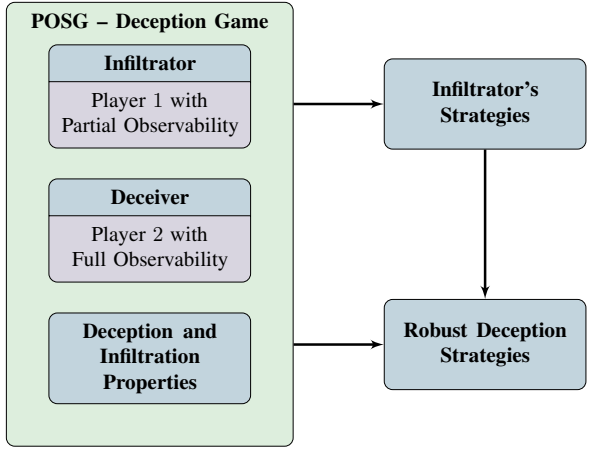
\begin{figure}[!t]
\centering
	\scalebox{0.8}{	%color scheme
%background 
\definecolor{bg}{HTML}{ddeedd}
\definecolor{comp}{HTML}{c2d4dd}
\definecolor{impl}{HTML}{b0aac0}
\centering
\begin{tikzpicture}[every node/.style={draw, text centered, shape=rectangle, rounded corners, text width=3cm, minimum height=1.5cm, inner sep=5pt}]
%\tikzstyle{outer}= [draw, text centered, shape=rectangle, text width=2cm, minimum height=1cm]
%\tikzstyle{inner}=[draw, text centered, shape=rectangle, rounded corners, text width=3.8cm, minimum height=1.1cm, inner sep=5pt]
\tikzset{
splitnode/.style={
       rectangle split,
       rectangle split parts=2,
       rectangle split part fill={comp,impl!50}
       }
}

\tikzset{
normalnode/.style={
	   fill=comp
       }
}

\tikzstyle{split}=[rectangle split,rectangle split parts=2]

\node[splitnode] (infiltrator) {\textbf{Infiltrator} \nodepart{second} Player~$1$ with Partial Observability};

\node[splitnode, below=.5cm of infiltrator] (deceiver) {\textbf{Deceiver} \nodepart{second} Player~$2$ with Full Observability};

\node[normalnode, below=.5cm of deceiver] (props) {\textbf{Deception and Infiltration Properties}};

\begin{scope}[on background layer]
	\node [fit = (current bounding box), inner sep = .7cm,  fill=bg] (posg) {};        
\end{scope}
	\node[above=-.4cm of infiltrator, draw=none, text width=4cm] (posglabel) {\textbf{POSG -- Deception Game}};

\node[normalnode, right=1.5cm of posg, yshift=2cm] (infiltration_strategies) {\textbf{Infiltrator's Strategies}};

\node[normalnode, right=1.5cm of posg, yshift=-2cm] (robust) {\textbf{Robust Deception Strategies}};

%%arrows
\draw ([yshift=2cm] posg.east) edge[-latex', very thick] ([] infiltration_strategies.west);
\draw ([yshift=-2cm] posg.east) edge[-latex', very thick] ([] robust.west);

\draw (infiltration_strategies) edge[-latex', very thick] (robust);

\end{tikzpicture}}
	\caption{The Robust Deception Problem.}
	\label{fig:flowchart_problem}
\end{figure}
}

The straightforward solution to Problem~1 can be given by solving a robust mixed-integer linear program (MILP) with variables as described next. 

For $s_{\circ}\in S_{\circ}$ and each action $\act\in\Act$, the  \emph{strategy variable} $\delta^{s_{\circ}}_{\act}\in\lbrace 0,1\rbrace$ denotes which action is active in each state $s_{\circ}$. If $\delta^{s_{\circ}}_{\act}=1$, then Player 1 takes action $\act$ in state $s_{\circ}$. 
For $s_{\circ}\in \spOne,$ and $s_{\Box} \in \spTwo $, the \emph{cost variables} $c_{s_{\circ}}\geq 0$ $c_{s_{\Box}}\geq 0$ and represent the expected cost of reaching $T\subseteq S$. $M$ is a large constant that automatically satisfies the constraints in~\eqref{eq:probcomputation_sg2_det} and \eqref{eq:probcomputation_sg3_det} if $\delta^{s_{\circ}}_{\act}=0$, which means the deceiver does not select action $\act$ in state $s_{\circ}$. $\gamma \in [0,1)$ is the discount factor to ensure that we have finite expected cost. 

We then have the following robust MILP that solves~Problem~1.
% For $s_{\circ}\in S_{\circ}$ and each action $\act\in\Act$, the  \emph{strategy variable} $\delta^{s_{\circ}}_{\act}\in[0,1]$ is the probability of choosing action $\act\in\Act$ upon observation $\obs$ for Player 1. For observation $\obs\in\ObsSym$ and each action $\act\in\Act$, the \emph{strategy variable} $\sched^{\obs}_{\act}\in[0,1]$ represents the probability of choosing action $\act\in\Act$ upon observation $\obs$ for Player 2.

	%The NLP reads as follows:
%		\begin{align}
%			\text{maximize} &\quad p_{\sinit}\label{eq:min_sg}\\
%			\text{subject to} &\nonumber \\
%			\forall s\in T.	 &\quad p_s=1\label{eq:targetprob_sg}\\
%						\forall \obs\in \ObsSym. &\quad \sum_{\act\in \Act}\sched^\obs_\act=1\label{eq:well-defined_probs_sg}\\
%												\forall s_{\circ}\in S_{\circ}. &\quad \sum_{\act\in \Act}\delta^{s_\circ}_\act=1\label{eq:well-defined_policy_sg}\\
%									\forall s_{\Box}\in S_{\Box}\setminus T.\,\forall \sched^{\ObsFun(s_{\Box})}_\act \in \sched_2	&\\ \quad p_{s_{\Box}} = \sum_{\act\in\Act}\sched^{\ObsFun(s_{\Box})}_\act  ~ \sum_{s'_{\circ}\in \S_{\circ}}	&\probmdp(s_{\Box},\act,s'_{\circ}) ~ p_{s'_{\circ}}\label{eq:probcomputation_sg1}\\
%				\forall s_{\circ}\in S_{\circ}\setminus T. \,\forall\act\in\Act.	&\\ \quad p_{s_{\circ}} =\sum_{\act\in \Act}\delta^{s_\circ}_\act \sum_{s'_{\Box}\in S_{\Box}}&\probmdp(s_{\circ},\act,s'_{\Box}) ~ p_{s'_{\Box}}\label{eq:probcomputation_sg2}
%		\end{align}	
		
 %\ra{????}

				\begin{align}
		&	\underset{c_{s_{\circ}},c_{s_{\Box}},\delta^{s_{\circ}}_{\act}}{\text{minimize}} \quad c_{\sinit}\label{eq:min_sg_det}\\
			&\text{subject to} \nonumber \\
		&	 c_s=0, \quad \forall s\in T&\label{eq:targetprob_sg_det}\\
				&	 \sum_{\act\in \Act}\sched^\obs_\act=1,\quad	\forall \obs\in \ObsSym&\label{eq:well-defined_probs_sg_det}\\
								&			 \sum_{\act\in \Act}\delta^{s_\circ}_\act=1,	\quad\forall s_{\circ}\in S_{\circ}\label{eq:well-defined_policy_sg_det}\\
								& c_{s_{\Box}} =  C(s_{\Box},\act)+\sum_{\act\in\Act}\sched^{\ObsFun(s_{\Box})}_\act  ~ \sum_{s'_{\circ}\in \S_{\circ}}	\probmdp(s_{\Box},\act,s'_{\circ}) ~ c_{s'_{\circ}},\nonumber\\
								&\qquad\qquad\qquad\qquad	\forall s_{\Box}\in S_{\Box}\setminus T,\,\forall \sched^{\ObsFun(s_{\Box})}_\act \in \sched_2 \label{eq:probcomputation_sg1_det}	\\ 
 & c_{s_{\circ}} \geq C(s_{\circ},\act)+ \gamma  ~ \sum_{s'_{\Box}\in S_{\Box}}\probmdp(s_{\circ},\act,s'_{\Box}) ~ c_{s'_{\Box}}-M ~(1-\delta^{s_{\circ}}_{\act}),\nonumber\\
 				&\qquad\qquad\qquad\qquad\forall s_{\circ}\in S_{\circ}\setminus T, \,\forall\act\in\Act\label{eq:probcomputation_sg2_det}	\\
							 & c_{s_{\circ}} \leq  C(s_{\circ},\act)+ \gamma  ~ \sum_{s'_{\Box}\in S_{\Box}}\probmdp(s_{\circ},\act,s'_{\Box}) ~ c_{s'_{\Box}}+M ~(1-\delta^{s_{\circ}}_{\act}),\nonumber\\
		&	\qquad\qquad\qquad\qquad \forall s_{\circ}\in S_{\circ}\setminus T, \,\forall\act\in\Act.	\label{eq:probcomputation_sg3_det}
		\end{align}	
		
The objective in~\eqref{eq:min_sg_det} minimizes the expected cost of the deceiver. We assign the expected cost of the states in the target set to 0 by the constraints in~\eqref{eq:targetprob_sg_det}. We ensure that the strategies of the infiltrator and the deceiver are well-defined with the constraints in~\eqref{eq:well-defined_probs_sg_det} and~\eqref{eq:well-defined_policy_sg_det}. The constraints in~\eqref{eq:probcomputation_sg1_det}--\eqref{eq:probcomputation_sg3_det} gives the computation for the expected cost in the states of the POSG.
\begin{proposition}
The robust MILP in \eqref{eq:min_sg_det} -- \eqref{eq:probcomputation_sg3_det} computes the \emph{maximal probability} of reaching $T$ under a (maximizing) randomized memoryless observation-based strategy.
\end{proposition}

Unfortunately, robust MILP's are notoriously hard to solve, except in special cases of small problems with few constraints~\cite{goetzmann2011optimization} or only having binary variables~\cite{bertsimas2003robust}. Recently,~\cite{pauphilet2016tractable} proposes a heuristic relaxation for robust MILP's using the affinely adjustable robust counterpart~\cite{ben2004adjustable}. However,  large robust MILP's, such as the robust MILP above for Problem 1, remain intractable in general.

%\mc{will look at this again}

\section{Tractable Approach}\label{sec:approach}

\begin{figure}[!t]
	\centering
	\scalebox{0.85}{%color scheme
%background 
\definecolor{bg}{HTML}{ddeedd}
\definecolor{comp}{HTML}{c2d4dd}
\definecolor{impl}{HTML}{b0aac0}
\centering
\begin{tikzpicture}[every node/.style={draw, text centered, shape=rectangle, rounded corners, text width=4.1cm, minimum height=1.5cm, inner sep=3pt}]
%\tikzstyle{outer}= [draw, text centered, shape=rectangle, text width=2cm, minimum height=1cm]
%\tikzstyle{inner}=[draw, text centered, shape=rectangle, rounded corners, text width=3.8cm, minimum height=1.1cm, inner sep=5pt]
\tikzset{
splitnode/.style={
       rectangle split,
       rectangle split parts=2,
       rectangle split part fill={comp,impl!50}
       }
}

\tikzset{
normalnode/.style={
	   fill=comp
       }
}

\tikzstyle{split}=[rectangle split,rectangle split parts=2]

%\node[splitnode] (deceiver) {\textbf{Deceiver} \nodepart{second} Player~$1$ with Full Observability};

%\node[splitnode, below=1.5cm of deceiver] (infiltrator) {\textbf{Infiltrator} \nodepart{second} Player~$2$ with Partial Observability};

\node [splitnode] (posg) {\textbf{Deception Game} \nodepart{second} One-sided POSG};        

\node[splitnode, right=1.0cm of posg] (pmdp) {\textbf{Symbolic Representation of Infiltration Strategies} \nodepart{second} Parametric Markov Decision Process};

\node[splitnode, below=2cm of pmdp] (strategy_set) {\textbf{Set of Infiltration Strategies} \nodepart{second} Strategies satisfying the Infiltration Objectives};

\node[splitnode, below=1.0cm of posg] (memory) {\textbf{Memory} \nodepart{second} Finite State Controller};

%%%arrows
%\draw ([yshift=1.59cm] posg.east) edge[-latex', very thick] ([] pmdp.west);
%\draw ([yshift=-1.59cm] posg.east) edge[-latex', very thick] ([] mdp.west);
%
\draw (posg) edge[-latex', very thick] (pmdp);
\draw (pmdp) edge[-latex', very thick] (strategy_set);
%
%\draw (spec) edge[-latex', very thick, dashed] (pmdp);
%\draw (spec) edge[-latex', very thick, dashed] (strategy);
%
\draw (memory) edge[-latex', very thick] (pmdp);
%%\draw (pmdp) edge[-latex', very thick] (memory);
%

\end{tikzpicture}}
	\caption{Computing Infiltration Strategies for Stage 1.}
	\label{fig:flowchart_infiltration}
\end{figure}
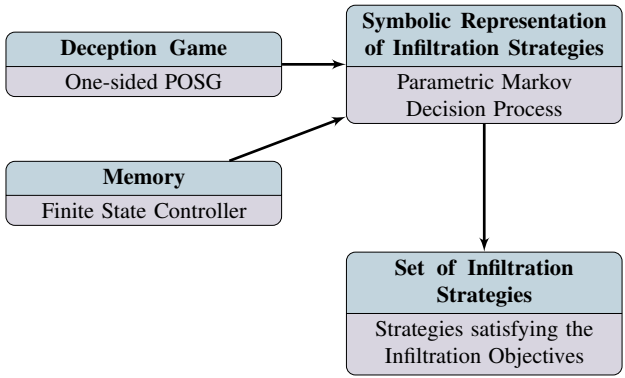

%\begin{figure*}[!t]
%	\centering
%%	\includegraphics[scale=0.4]{pics/flowchart_general}
%	\scalebox{0.85}{\input{pics/flowchart_robustification}}
%	\caption{Computing a Robust Deception Strategy}
%	\label{fig:flowchart_robustification}
%\end{figure*}

%\subsection{Technical Background}\ra{This should go to the beginning of the paper}
In this section, we propose a two-stage approach for solving Problem 1. We begin by giving some key formalisms.

\subsubsection{Parametric MDPs}
Instead of having fixed probabilities in the transition function of an MDP, we allow to describe the transition probabilities of an MDP as polynomials over a fixed set of variables. By having different values for the parameters, we can induce different MDPs.
\begin{definition}[pMDP]
  \label{def:pmdp}
  A \emph{parametric Markov decision process} (pMDP) $\pmdp$ is a tuple $\pMdpInit$ with a finite
  (or countably infinite) set $\states$ of \emph{states}, \emph{initial state} $\sinit\in\states$, a finite set
  $\Act$ of \emph{actions}, a finite set $\Var$ of parameters, and a \emph{transition function}
  $\probmdp\colon \states\times\Act\times\states\rightarrow \poly[\Var]$.
\end{definition}

%\nj{add reward to definition}\sj{Rewards are unchanged, but are not yet part of the MDP def}

Applying an \emph{instantiation} $u\colon\Var\to\R$ to a pMDP $\mdp$, denoted $\mdp[u]$, replaces each polynomial
$ \poly[\Var]$ in $\mdp$ by $ \poly[u]$. $\mdp[u]$ is also called the \emph{instantiation} of $\mdp$ at $u$.
Instantiation $u$ is \emph{well-defined} for $\mdp$ if the replacement yields probability distributions, \ie,
if $\mdp[u]$ is an MDP.

\begin{definition}[pMDP Synthesis Problem]
	Given a pMDP $M$ and a property $\varphi$, the synthesis problem is to compute an instantiation $u$ such that $M[u] \models \varphi$, if one exists.
	\end{definition}

The pMDP synthesis problem amounts to solving the following NLP.

		\begin{align}
		&	\underset{c_s,V}{\text{maximize}} \quad c_{\sinit}\label{eq:min_mdp}\\
		&	\text{subject to} \nonumber \\
							 & c_{\sinit}\leq \kappa,\label{eq:strategyah:lambda}\\
		 & c_s=0,\quad	\forall s\in T	\label{eq:targetprob_mdp}\\
				 & \probmdp(s,\act,s')\geq 0,\quad \forall s,s'\in S,\, \forall\act\in\Act\label{eq:well-defined_probs_mdp}\\
					 & \sum_{s'\in S}\probmdp(s,\act,s')=1,\quad	\forall s\in S,\, \forall\act\in\Act	\label{eq:well-defined_probs_mdp1}\\
%%&\quad  \lambda \geq p_{\sinit}\label{eq:probthreshold_mdp}\\
								& 	 c_s \geq C(s,\act) + \gamma  \sum_{s'\in S}	\probmdp(s,\act,s')~ c_{s'},\label{eq:probcomputation_mdp}\\
								&\qquad\qquad \qquad\qquad\forall s\in S\setminus T,\,\forall \act\in\Act.\nonumber\end{align}
For $s \in S$, the \emph{cost variable} $c_s\geq 0$ represents an upper bound of expected cost of reaching target set $T\subseteq S$, and the \emph{parameters} in set $V$ enter the NLP as part of the functions from $\mathbb{Q}[V]$ in the transition function $\probmdp$.
	
One particular efficient method for finding parameter instantiations is given in~\cite{cubuktepe-et-al-qcqp-techreport}, which solves the NLP in~\eqref{eq:min_mdp}--\eqref{eq:probcomputation_mdp} via a reformulation to a convex-concave programming problem~\cite{lipp2016variations} and an efficient integration of model checking calls to improve its performance.
%Proving the absence of such an instantiation is harder, a particular efficient method relying on model checking methods is given in \cite{QDJJK16}.
We are now ready to outline the proposed tractable approach.

\subsection{Stage 1: Computing Infiltration Strategies}
In this section, we are given a POSG and are interested in computing several sufficiently strong infiltrator strategies. For an overview of the approach in Stage 1, see Figure~\ref{fig:flowchart_infiltration}.

\begin{problem}
	Given a POSG $\posg$, compute several sufficiently strong strategies for the infiltrator. 
\end{problem}

We detail the approach first for an infiltrator that uses memoryless strategies. 	We then generalize the approach to any (finite but fixed) amount of memory for the infiltrator.
The approach extends the reduction from POMDPs to pMCs in~\cite{junges2018pomdppmc}.

\subsubsection{Memoryless strategies}
A memoryless strategy maps observations to distributions over actions.
Equivalently, such strategy maps observation-action pairs to probabilities (ObAct-probabilities).
Any memoryless strategy is uniquely defined by its ObAct-probabilities. 
Instead of finding suitable strategies for the infiltrator, we can thus reformulate our approach to finding suitable ObAct-probabilities. 
Hence, a suitable strategy can be described as parameter values that satisfy the property of an pMDP.
To find the ObAct-probabilities, we construct an equivalent pMDP as illustrated in the following example.

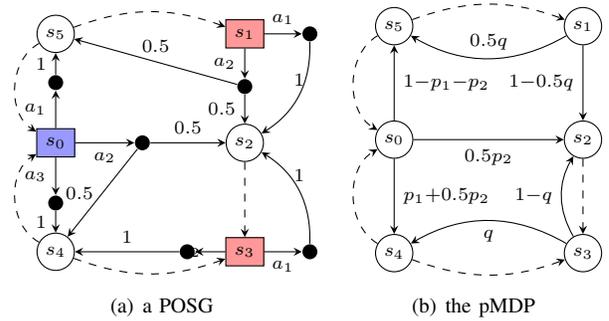
\begin{figure}
\subfigure[a POSG]{
	\begin{tikzpicture}[font=\scriptsize]
\node[rectangle, draw, fill=blue!40] (s0) {$s_0$};
\node[state, right=2cm of s0, fill=white!40] (s2)  {$s_2$};
\node[rectangle, draw, above=of s2, fill=red!40] (s1)  {$s_1$};
\node[rectangle, draw, below=of s2, fill=red!40] (s3)  {$s_3$};
\node[state, draw, below=of s0] (s4) {$s_4$};
\node[state, draw, above=of s0] (s5) {$s_5$};

\node[circle, inner sep=2pt, fill=black, above=0.5cm of s0] (a1) {};
\node[circle, inner sep=2pt, fill=black, left=1cm of s2] (a2) {};
\node[circle, inner sep=2pt, fill=black, below=0.5cm of s0] (a3) {};
\node[circle, inner sep=2pt, fill=black, right=0.5cm of s1] (a4) {};
\node[circle, inner sep=2pt, fill=black, below=0.4cm of s1] (a5) {};
\node[circle, inner sep=2pt, fill=black, right=0.5cm of s3] (a6) {};
\node[circle, inner sep=2pt, fill=black, left=0.4cm of s3] (a7) {};

\draw[->] (s0) -- node[left] {$\act_1$} (a1);
\draw[->] (s0) -- node[below] {$\act_2$} (a2);
\draw[->] (s0) -- node[left] {$\act_3$} (a3);
\draw[->] (s1) -- node[above] {$\act_1$} (a4);
\draw[->] (s1) -- node[left] {$\act_2$} (a5);
\draw[->] (s3) -- node[below] {$\act_1$} (a6);
\draw[->] (s3) -- node[left] {$\act_2$} (a7);

\draw[->] (a1) -- node[left] {$1$} (s5);
\draw[->] (a2) -- node[above] {$0.5$} (s2);
\draw[->] (a2) -- node[left] {$0.5$} (s4);
\draw[->] (a3) -- node[left] {$1$} (s4);

\draw[->] (a4) edge[bend left] node[above] {$1$} (s2);

\draw[->] (a5) -- node[above] {$0.5$} (s5);
\draw[->] (a5) -- node[left] {$0.5$} (s2);

\draw[->] (s2) edge[dashed] (s3);
\draw[->] (s4) edge[dashed, bend right=20] (s3);
\draw[->] (s4) edge[dashed, bend left=60] (s0);

\draw[->] (s5) edge[dashed, bend left=20] (s1);
\draw[->] (s5) edge[dashed, bend right=60] (s0);

\draw[->] (a6) edge[bend right] node[above] {$1$} (s2);

\draw[->] (a7) edge[] node[above] {$1$} (s4);
\end{tikzpicture}% 
\label{fig:posgpmdp:posg} 
}
\subfigure[the pMDP]{
	\begin{tikzpicture}[font=\scriptsize]
\node[state, draw] (s0) {$s_0$};
\node[state, right=2cm of s0] (s2)  {$s_2$};
\node[state, draw, above=of s2] (s1)  {$s_1$};
\node[state, draw, below=of s2] (s3)  {$s_3$};
\node[state, draw, below=of s0] (s4) {$s_4$};
\node[state, draw, above=of s0] (s5) {$s_5$};

\draw[->] (s1) edge[bend left] node[above] {$0.5   q$} (s5);
\draw[->] (s1) edge node[left] {$1{-}0.5   q$} (s2);
\draw[->] (s3) edge[bend left] node[left] {$1{-}q$} (s2);
\draw[->] (s3) edge[bend right] node[below] {$q$} (s4);
\draw[->] (s0) edge node[right] {$1{-}p_1{-}p_2$} (s5);
\draw[->] (s0) edge node[right] {$p_1{+}0.5   p_2$} (s4);
\draw[->] (s0) edge node[below] {$0.5   p_2$} (s2);

\draw[->] (s2) edge[dashed] (s3);
\draw[->] (s4) edge[dashed, bend right=20] (s3);
\draw[->] (s4) edge[dashed, bend left=60] (s0);

\draw[->] (s5) edge[dashed, bend left=20] (s1);
\draw[->] (s5) edge[dashed, bend right=60] (s0);

\end{tikzpicture}% 
\label{fig:posgpmdp:pmdp}
}
\caption{From POSG to pMDP}
\end{figure} 
 
\begin{example}
	Consider the POSG in Fig.~\ref{fig:posgpmdp:posg}. 
	The states for Player 1 are given by circles. 
	Similarly, the rectangles depict the states for Player 2. We indicate the actions from the states with solid lines to black dots. The corresponding probabilities from the black dots give the probability of transitioning to the next state. 
	%In the figure, all actions have Dirac-distributions. 
	Thus, we draw dashed (direct) lines from the Player 1 (Player 2) nodes.
	The colors of states indicate the corresponding observations.
	
	The corresponding pMDP for a memoryless, observation-based strategy for player 2 is given in Fig.~\ref{fig:posgpmdp:pmdp}. The actions of player 1 are unchanged. As there is no more nondeterminism in the states of player 2, we can also view these states as player 1 states. Moreover, to avoid clutter, we omit action indications.
	Consider state $s_3$ with outgoing action $\act_2$, which leads with probability $1$ to $s_4$. Under a strategy, we take this action with probability $q$: Thus, in the pMDP there is an arc with probability $q$ from $s_3$ to $s_4$.
	If we take action $\act_2$ with probability $q$, we take action $\act_1$ with probability $1{-}q$. 
	The same probabilities for taking the actions have to hold in state $s_1$. Observe that the probabilities are not immediately reflected, as there are two paths from $s_1$ to $s_2$.
	The parameters in $s_0$ are different. Indeed, there are $2$ parameters as there are three actions to take.
\end{example}
We formalize the construction below for situations, where the infiltrator has three available actions in each state, to simplify the notation.
\begin{definition}
	Given a POSG $\PosgInit$ with SG $\sgInit$. Let $\forall s \in \spTwo~|\Act(s)| = 3$. Without loss of generality, refer to $\Act(s) = \{ \act_1, \act_2, \act_3 \}$ if $s \in \spTwo$.
	The pMDP $\pMdpInit[']$ is the \emph{corresponding pMDP for POSG $\posg$}, if $S' = \spOne \cup \spTwo$, $\sinit' = \sinit$, $\Act' = \Act \cup \{ \bot \}$, $\Var = \{ p^z_1, p^z_2 \mid z \in \ObsSym \}$ and  $\probmdp'(s,a,s') =$ \[ \begin{cases} \probmdp(s,\act,s') & \text { if } s \in \spOne \\
		\sum_{i=1}^3 f^z_i   \probmdp(s,\act_i,s') & \text { if } s \in \spTwo, \act = \bot, z = \ObsFun(s) \\
		0 & \text{otherwise.}
 \end{cases} 
 \]
 where $f^z_i$ denotes $p_1^z$, $p_2^z$ and $1-p_1^z-p_2^z$, respectively.
\end{definition}
The construction is straightforward to adapt to a varying number of actions per state: 
Indeed, we just need additional variables $p^z_i$ indicating that we take action $i$ in a state with observation $z$.

We denote the memoryless strategy $\IStrat(u)$ for the infiltrator in the deception game that is induced by a valuation $u$ in the infiltration-pMDP.

\begin{theorem}
Given a deception game $\posg$, with its corresponding infiltration-pMDP $\infiltration(\posg)$, and a property $\varphi$. 
Let $u$ be a solution for the pMDP synthesis problem for  the property $\varphi$. 
Then $\IStrat(u)$ is a sufficiently strong strategy for the infiltrator on the $G$ and $\varphi$.
\end{theorem}

%%As indicated before, and visualized in Fig.~\ref{fig:infiltrationstratscheme}, we compute a selection of these strategies by partitioning the search space. 
%%We partition the strategies by putting constraints on the randomization of a strategy. 
%%As example, we add a constraint that the infiltrator takes action $\act$ with probability $<0.5$ or $\geq 0.5$. 
%%Such constraints are naturally expressible as additional linear constraints over the parameters of the infiltration pMDP, and can be immediately added to the encoding.

\subsubsection{Adding Memory to the Infiltrator Strategy}
The ideas for the synthesis of memoryless strategies can be lifted to a finite ($k$) memory setting. 
We apply an $k-$unfolding of the POSG, and then search for a memoryless infiltrator strategy.
The idea is to create $k$ copies of the POSG, and allow the infiltrator to freely switch between copies. 
The different copies correspond to the internal memory of the infiltrator.
%\begin{example}
%
%\end{example}
%The infiltrator can obviously observe in which the POSG copy is active, as the infiltrator can freely switch between the copies. 

%We assume that the deceiver has full observation. In particular, this means that the deceiver observes the internal memory state of the infiltration strategy. 
%This assumption is necessary to stay in the realm of standard pMDP solvers. \mc{maybe say something about why this assumption is necessary?}
%Thus, in this step, the computation of the infiltration strategies over-estimates the capabilities of the deceiver. 
%Put differently, the computed attacks are also strong against an overly strong deceiver.

\subsection{Stage 2: Computing a Robust Deception Strategy}
In this section, we assume that the deceiver obtained $N$ strategies for the infiltrator. 
Then, the task of the deceiver is to compute a strategy that minimizes the worst-case expected cost $\psi=\expRewProp{\kappa}{T}$ to reach a target set $T$ against any of the $N$ infiltration strategies under all deceiver policies. 
%We distinguish two scenarios, corresponding to the following two variants of the problem statement.
%\emph{The difference lies in the information available to the deceiver}:
%If the deceiver strategy does depend on the internal memory node of an infiltrator, but only on the POSG state $s$, we say that the infiltrator memory is \emph{opaque}. 
%The strategy for the deceiver has then to be computed under partial observability. 
%Thus, the application of the strategy yields again a partial observable system.
%\begin{problem}
%	Given a deception game, a set of opaque infiltration strategies, compute a deception strategy that is sufficiently strong for the set of infiltration strategies.
%\end{problem}
If the deception strategy has access to the memory node of the infiltration strategy, we call the infiltration memory \emph{transparent}.
\begin{problem}
	Given a deception game, a set of transparent infiltration strategies, compute a deception strategy that is optimal under all deception strategies against the set of infiltration strategies.
\end{problem}

\subsubsection{Deceiving against Transparent-Memory Infiltrator Strategies}
We focus on memoryless infiltration strategies.
Memoryless infiltration strategies are the most prominent transparent memory case, and for transparent memory, each finite memory strategy can be reduced to a memoryless strategy by unfolding the memory into the MDP, as in Stage~1. 

To compute a robust deception strategy against transparent infiltrator strategies, we remove the uncertain constraints of the robust MILP in~\eqref{eq:min_sg_det} -- \eqref{eq:probcomputation_sg3_det} by using the infiltrator strategies that we obtained in Stage~1. Then, we reduce the problem of synthesizing a robust deception strategy problem to solving an MILP instead of solving a robust MILP. We now give the details of the resulting MILP with the infiltrator strategies.

We define the following variables for the following MILP:
For $s_{\circ}\in \spOne$, $s_{\Box} \in \spTwo $, and for each infiltrator strategy $\sched^{i,\ObsFun(s_{\Box})}, i=1,\ldots,N$, the \emph{cost variables} $ ~c^i_{s_{\circ}}\geq 0$ and $ ~c^i_{s_{\Box}}\geq 0$ give the expected cost of reaching $T\subseteq S$ for each infiltrator strategies.
 For $s_{\circ}\in S_{\circ}$ and each action $\act\in\Act$, the  \emph{strategy variable} $\delta^{s_{\circ}}_{\act}\in\lbrace 0,1\rbrace$ denotes which action is active in each state $s_{\circ}$. If $\delta^{s_{\circ}}_{\act}=1$, then the deceiver takes action $\act$ in state $s_{\circ}$.

For observation $\obs\in\ObsSym$, for each action $\act\in\Act$ and for each infiltrator strategy $\sched^{i,\ObsFun(s_{\Box})}, i=1,\ldots,N$, $\sched^{i,\ObsFun(s_{\Box})}_\act$ represents the probability of choosing action $\act\in\Act$ upon observation $\obs$ for each strategy of Player 2. Similar to the Problem~1, $M$ is a large constant that automatically satisfies the constraints in~\eqref{eq:probcomputation1_milp_det} and \eqref{eq:probcomputation2_milp_det} if $\delta^{s_{\circ}}_{\act}=0$, which means the deceiver does not select action $\act$ in state $s_{\circ}$. 

We thus have the following MILP:

				\begin{align}
			&\underset{ c^i_{s_{\circ}}, c^i_{s_{\Box}},\delta^{s_{\circ}}_{\act}}{\text{minimize}} \;\;\underset{i}{\text{max}}\quad r^i_{\sinit}\label{eq:min_milp_det}\\
			&\text{subject to} \nonumber \\
		&	 ~c^i_s=0, \quad\forall s\in T,~i=1,\ldots,N	 \label{eq:targetprob_milp_det}\\
											&	\sum_{\act\in \Act}\delta^{s_\circ}_\act=1, \quad\forall s_{\circ}\in S_{\circ}\label{eq:well-defined_policy_milp_det}\\
									&  ~c^i_{s_{\Box}} = C(s_{\Box},\act)+ \sum_{\act\in\Act}\sched^{i,\ObsFun(s_{\Box})}_\act   \sum_{s'_{\circ}\in \S_{\circ}}	\probmdp(s_{\Box},\act,s'_{\circ})   ~c^i_{s'_{\circ}},\nonumber\\
																	&	\qquad\qquad\qquad\qquad\forall s_{\Box}\in S_{\Box}\setminus T,~i=1,\ldots,N \label{eq:probcomputation_milp_det}\\
				&  ~c^i_{s_{\circ}} \geq C(s_{\circ},\act)+ \gamma  \sum_{s'_{\Box}\in S_{\Box}}\probmdp(s_{\circ},\act,s'_{\Box})   ~c^i_{s'_{\Box}}-M (1-\delta^{s_{\circ}}_{\act}),\nonumber\\
				&\qquad\qquad\qquad\qquad	\forall s_{\circ}\in S_{\circ}\setminus T, \,\forall\act\in\Act,~ i=1,\ldots,N	\label{eq:probcomputation1_milp_det}\\
								&  ~c^i_{s_{\circ}} \leq C(s_{\circ},\act)+ \gamma   \sum_{s'_{\Box}\in S_{\Box}}\probmdp(s_{\circ},\act,s'_{\Box})   ~c^i_{s'_{\Box}}+M (1-\delta^{s_{\circ}}_{\act}),\nonumber\\
							&	\qquad\qquad\qquad\qquad\forall s_{\circ}\in S_{\circ}\setminus T, \,\forall\act\in\Act,~i=1,\ldots,N.	 \label{eq:probcomputation2_milp_det} 
		\end{align}	
		
 The objective in~\eqref{eq:min_milp_det} minimize the worst-case expected cost of deception against the infiltrator strategies. The constraint in~\eqref{eq:targetprob_milp_det} sets the expected cost of the states in $T$ to be 0. Constraint~\eqref{eq:well-defined_policy_milp_det} ensures that the deceiver picks one of the actions in $\Act$ for each state in $S_{\circ}$. The constraints~\eqref{eq:probcomputation_milp_det}--\eqref{eq:probcomputation2_milp_det} concerns the cost computation for each state in $\spOne$, and $\spTwo$.

By solving the MILP in~\eqref{eq:min_milp_det} -- \eqref{eq:probcomputation2_milp_det}, we construct a strategy for the deceiver that is sufficiently strong against infiltrator strategies. By construction, we have the following theorem.

\begin{theorem}
The MILP in~\eqref{eq:min_milp_det} -- \eqref{eq:probcomputation2_milp_det} computes a sufficiently strong strategy for the deceiver against $N$ strategies of the infiltrator.
\end{theorem}

\begin{table*}[t]\label{table:actionsvscosts}
\centering
\begin{tabular}{|*5{p{27mm}|}}
\hline
\multicolumn{2}{|c|}{States}&\multicolumn{2}{c|}{Actions}&\multicolumn{1}{c|}{loss ($l$) }\\
\cline{1-4}
Infiltrator's Position ($x_A$)& Detection mode ($d$)&Infiltrator ($a_I$)&Deceiver ($a_D$)&\\
\hline
$any$ & false& compromise & -- & $-2$\\
\cline{1-2}
\hline
$\mathrm{layer}_i$ & false& exfiltrate & -- & $15i$ \\
\hline
$\mathrm{layer}_i$&false&takedown&--&$25i$\\
\hline
$any$&true&compromise&engage&$-4$\\
\hline
$\mathrm{layer}_i$&true&exfiltrate&engage&$-2$\\
\hline
$\mathrm{layer}_i$&true&takedown&engage&$25i$\\
\hline
$any$&true&compromise&block&$-2$\\
\hline
$any$&true&exfiltrate&block&$0$\\
\hline
$any$&true&takedown&block&$0$\\
\hline
\end{tabular}
\caption{The losses incurred when both players take simultaneous actions at different states.}
\end{table*}

\section{Numerical Experiments}\label{sec:experiments}

At this point, we are ready to return to the motivating example. We define the expected discounted cost for the deceiver corresponding to each action the infiltrator takes by
$$
L =  \sum_{t=1}^\infty \gamma^{t-1}\cdot l^{(t)},
$$
where $\gamma$ is the discount factor and $l^{(t)}$ denotes the loss at stage $t$. For this example, the specification of the deceiver is $\psi = \er_{\min}$, which is to minimize $L$ against all infiltration strategies. Table~\ref{table:actionsvscosts} outlines the elements of the one-sided deception POSG. Note that since the players take actions concurrently, the costs depend on their joint actions. Furthermore, some of the costs are dependent on the layer index $i$ the infiltrator has accessed. For more details about the example, the interested reader is referred to~\cite{horak2017manipulating}.

We first describe our results on a 4-layer network, given by~\cite{horak2017manipulating}. Then, we demonstrate our approach on a 12-layer network. For each network, we first compute  infiltration strategies for the deceiver using the approach in Stage 1. Then, using these infiltration strategies, we compute a robust deceiver strategy, which is given by in Stage 2. The arising pMDP problems from Stage 1 is solved using the approach in~\cite{cubuktepe-et-al-qcqp-techreport}. We solve the MILP problems from Stage 2 using the MILP solver GUROBI~\cite{gurobi}.

For the 4-layer network, we first construct the one-sided deception POSG with 2 memory nodes. The POSG consists of 49 states, 8 action choices for the deceiver, and 34 actions for the infiltrator. After computing an optimal deceiver strategy, the worst-case induced cost against $1000$ infiltration strategies obtained from the approach in Stage 1 is 282.22. The obtained cost is comparable to the approach given by~\cite{horak2017manipulating}. The procedure for obtaining the infiltration strategies took 193.29 seconds, and time to compute the optimal deceiver strategy is 216.92 seconds.

The optimal deceiver strategy that we obtained is to engage the infiltrator in first 2 layers, then blocking the infiltrator in the last layers. This seems to be beneficial compared to always blocking the infiltrator, which leads to a worst-case expected cost of 341.72, and always engaging the infiltrator, which leads to a worst-case expected cost of 304.05.

We also give the results with different number of infiltrator strategies on a 4-layer and 12-layer network in Table~\ref{table:actionsvscosts2}. The expected cost of the defender increases with increasing number of infiltrator strategies in all cases. Also, by increasing number of layers, the deceiver can craft an optimal strategy in a 12-layer network that incurs a less expected loss compared to the optimal strategy in a 4-layer network. However, if the deceiver always blocks or engages, the expected cost increases with increasing number of layers.

\begin{table}[t]\label{table:actionsvscosts2}
\centering
\begin{tabular}{|l|l|l|l|}
\hline
Number of infiltrator strategies        & \multirow{2}{*}{10} & \multirow{2}{*}{100} & \multirow{2}{*}{1000} \\ \cline{1-1}
Number of Layers \& deceiver strategies &                     &                      &                       \\ \hline
4 \& Always engage                      & 260.17              & 297.45               & 304.05                \\ \hline
4 \& Always block                       & 246.12              & 286.19               & 341.72                \\ \hline
4 \& Optimal                            & 228.64              & 256.81               & 282.22                \\ \hline
12 \& Always engage                     & 277.71              & 301.46               & 313.91                   \\ \hline
12 \& Always block                      & 256.39              & 294.65               & 308.12                   \\ \hline
12 \& Optimal                           & 217.76              & 239.71               & 261.49                   \\ \hline
\end{tabular}
\caption{Expected loss of the defender in various scenarios.}
\end{table}

\section{Conclusion}
\label{sec:conclusion}

We presented an approach to solve a partially observable stochastic game (POSG), where one of the player has full observability over the states, and the other player only has partial observability. We formulated the problem as a robust mixed-integer linear program, which is intractable to solve in general. To obtain a more scalable approach, we computed a robust optimal strategy for the deceiver by synthesizing a set of infiltration strategies using parameter synthesis in parametric Markov decision processes. Using a mixed-integer linear program and the infiltration strategies, we computed the robust deception strategy. We illustrated our approach on a POSG model for network security and we showed that we can handle larger networks compared to the previous approaches in the literature.

Future work concerns removing the transparency approach, which means the deceiver has access to the memory node of the infiltration strategy, if the infiltration strategy is memory-based. Also, we will explore some of the recent methods to solve mixed-integer linear problems proposed in the literature, such as methods proposed in~\cite{dutta2017output,dutta2018learning}.

\bibliographystyle{plain}
\bibliography{literature}

\end{document}